%% file: eacl2023.tex
\pdfoutput=1
\documentclass[11pt]{article}

\usepackage{EACL2023}
\usepackage{times}
\usepackage{latexsym}

\usepackage[T1]{fontenc}
\usepackage[utf8]{inputenc}

\usepackage{microtype}

\usepackage{inconsolata}
\usepackage{multirow}
\usepackage{amsfonts}
\usepackage{booktabs, siunitx}
\usepackage{pdfpages}
\usepackage{graphicx}
\usepackage{times}
\usepackage{microtype}
\usepackage{latexsym}
\usepackage{pifont}
\usepackage[T1]{fontenc}    
\usepackage[utf8]{inputenc}
\usepackage[inline]{enumitem}
\usepackage{pythonhighlight}
\usepackage{amsmath}
\usepackage{amssymb}

\newcommand{\code}[1]{{\ttfamily#1}}

\title{Generative Knowledge Selection for Knowledge-Grounded Dialogues}

\author{Weiwei Sun,\quad Pengjie Ren,\quad Zhaochun Ren\thanks{~~Corresponding author.} \\
       Shandong University, Qingdao, China \\ \texttt{sunnweiwei@gmail.com},\quad\texttt{\{renpengjie,zhaochun.ren\}@sdu.edu.cn}
}

\begin{document}
\maketitle
\begin{abstract}
% Knowledge-grounded dialogue
% Large language 
Knowledge selection is the key in knowledge-grounded dialogues (KGD), which aims to select an appropriate knowledge snippet to be used in the utterance based on dialogue history. 
Previous studies mainly employ the classification approach to classify each candidate snippet as ``relevant'' or ``irrelevant'' independently.
However, such approaches neglect the interactions between snippets, leading to difficulties in inferring the meaning of snippets.
Moreover, they lack modeling of the discourse structure of dialogue-knowledge interactions.
We propose a simple yet effective generative approach for knowledge selection, called \textsc{GenKS}.
\textsc{GenKS} learns to select snippets by generating their identifiers with a sequence-to-sequence model.
\textsc{GenKS} therefore captures intra-knowledge interaction inherently through attention mechanisms.
Meanwhile, we devise a \emph{hyperlink} mechanism to model the dialogue-knowledge interactions explicitly.
We conduct experiments on three benchmark datasets, and verify \textsc{GenKS} achieves the best results on both knowledge selection and response generation.
\end{abstract}

\input{sections/01-Introduction}
\input{sections/02-RelatedWork}

\input{sections/03-Method}

\input{sections/04-Experiment}
\input{sections/05-Results}
\input{sections/06-Conclusion}

% \newpage

\bibliography{anthology,custom}
\bibliographystyle{acl_natbib}

\newpage

\input{sections/07-Appendix}
\end{document}

%% file: sections/01-Introduction.tex
\section{Introduction}
To improve the informativeness in open-domain dialogue agents~\citep{DeFreitas2020TowardsAH}, knowledge-grounded dialogues (KGD) are proposed to leverage external structured~\citep{Liu2019KnowledgeAC} and unstructured~\citep{Dinan2019WizardOW} knowledge to dialogue responses. 
In KGD, it is pivotal to embed factual and conversationally appropriate knowledge in responses.
Two classes of approaches are considered to embed knowledge: \emph{end-to-end} and \emph{pipeline}.
End-to-end models, such as FiD~\citep{Izacard2021LeveragingPR}, process the document and generate the response in one shot.
However, they tend to misuse knowledge~\citep{Adolphs2021ReasonFT}.
Pipeline models address this problem by explicitly identifying a specific knowledge snippet to be used in the response~\citep{Adolphs2021ReasonFT}.
Typically, pipeline KGD approaches have two sub-steps, i.e., knowledge selection and response generation~\citep{Dinan2019WizardOW,Kim2020SequentialLK}:
The former aims to select knowledge snippets from passages, and the latter generates responses based on them.
Knowledge selection plays a vital role in KGD as it directly determines the content of the response~\citep{Lian2019LearningTS,Meng2020DukeNetAD}.
In this paper, we focus on selecting knowledge snippets for dialogue to enhance pipeline KGD models.

\begin{figure}[t]
 \centering
 \includegraphics[width=1\columnwidth]{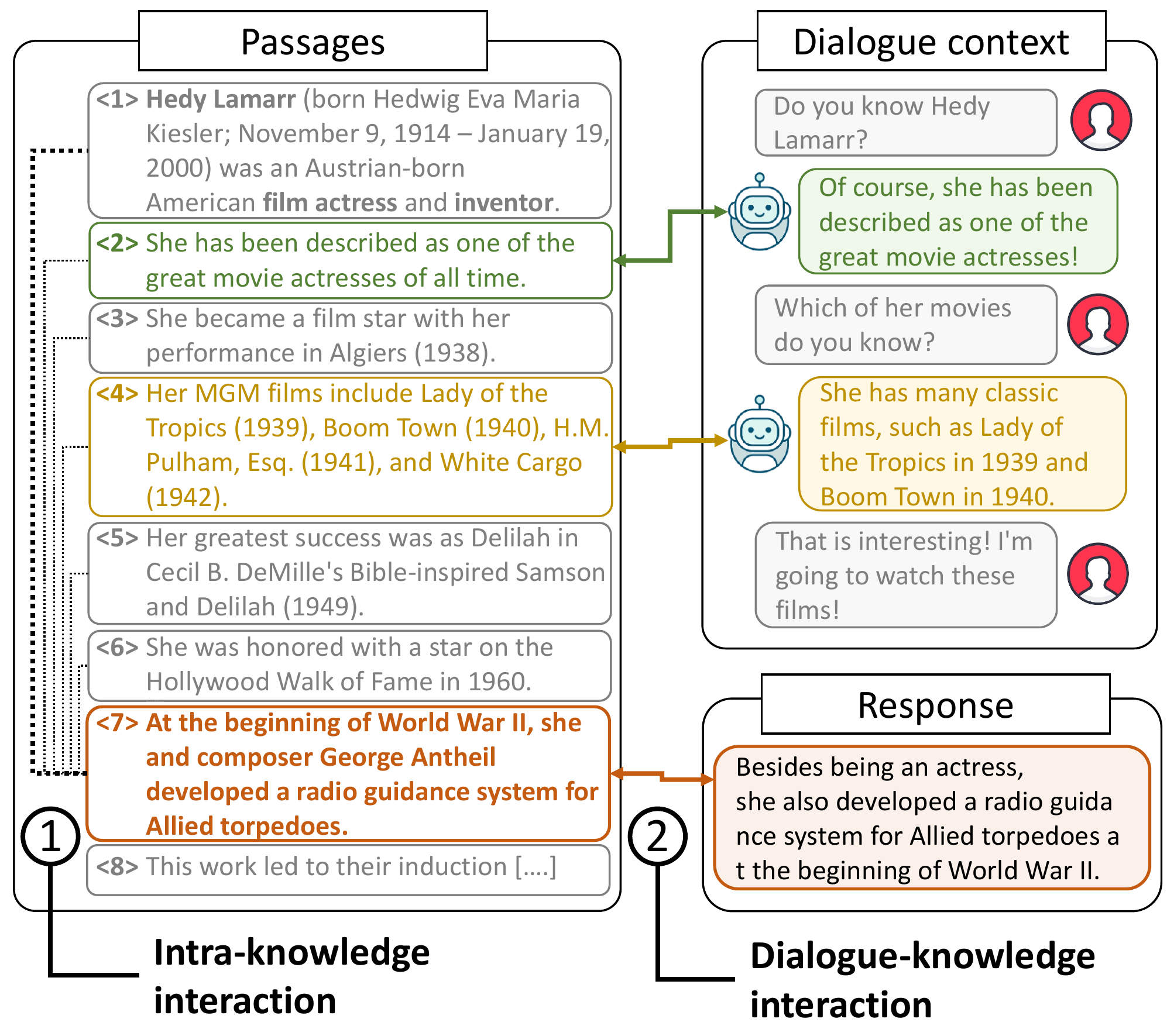}
 \caption{An example of knowledge-grounded dialogues. The dialogue agent selects a knowledge snippet (i.e., \code{<7>}) from passages and generates a response based on it. Intra-knowledge interactions and dialogue-knowledge interactions are denoted by \ding{192} and \ding{193}, respectively.
 }
\label{fig:motivation}
\end{figure}
The \emph{Classification} paradigm dominates knowledge selection studies. 
In this paradigm, each snippet is independently classified as ``relevant'' or ``irrelevant''~\citep{Dinan2019WizardOW,Zhao2020KnowledgeGroundedDG}.
However, these approaches ignore \emph{knowledge interactions}, which refer to flows of information within the knowledge or between knowledge and dialogues.
As shown in Figure \ref{fig:motivation}, we identify two types of knowledge interactions in KGD:
\begin{description}[leftmargin=10pt,itemsep=0pt, topsep=0pt]
    \item[Intra-knowledge interaction] 
    Intra-knowledge interaction refers to the interactions between snippets.
    It is worth noting that the meaning of a knowledge snippet is context-dependent and can be ambiguous when taken individually. 
    For example, the \code{<8>} snippet in Figure \ref{fig:motivation} \emph{``This work led to their''} has a referential element \emph{their}, and is difficult to identify its meaning without knowing the remaining context of the sentence.
    However, with the existence of the remaining context, we can quickly infer that it refers to \emph{Lamarr and George Antheil}.
    \emph{This problem challenges existing methods when selecting knowledge on new topics.}
    
    \item[Dialogue-Knowledge interaction]
    Previous works also neglect interactions between dialogue and knowledge.
    There is a discourse structure and smooth transition of involved knowledge in multi-turn dialogue.
    For example, \emph{Lamarr}'s profession mentioned in the dialogue in Figure~\ref{fig:motivation} is demonstrated in a parallel and multi-perspective manner, while some other cases follow a shallow-to-deep structure in dialogue.
    
\end{description}
Some recent efforts attempt to fix these problems within the classification paradigm; for example, \citet{Li2022EnhancingKS} build a semantic graph for passages to capture intra-knowledge interaction, \citet{Kim2020SequentialLK} propose sequential knowledge selection to model the dialogue-knowledge interaction as latent variables.
However, they are complicated, lack deep semantic interactions, and are challenging to model the two types of knowledge interaction simultaneously.

In this work, we propose \textbf{\textsc{GenKS}} (\textbf{Gen}erative \textbf{K}nowledge \textbf{S}election), a simple yet effective \emph{generative} model that addresses these challenges.
\textsc{GenKS} first assigns an identifier to each snippet, feeds all the snippets into the model simultaneously, and then selects snippets by generating their identifiers with a sequence-to-sequence Transformer model (e.g., BART~\cite{Lewis2020BARTDS}).
Compared with KGD methods with the \emph{classification} paradigm, 
\textsc{GenKS} captures interactions between knowledge snippets through the \emph{self-attention} mechanism in Transformer~\citep{Vaswani2017AttentionIA}.
Therefore, \textsc{GenKS} can obviate the ambiguity in snippets with the existence of the rest context and improve the understanding of knowledge. 
Moreover, we propose a \emph{hyperlink} method to capture the dialogue-knowledge interactions explicitly and effectively.
Finally, we propose to joint knowledge selection and response generation within one generative model.

We evaluate our proposed method on three public KGD datasets: Wizard of Wikipedia~\citep{Dinan2019WizardOW}, Holl-E~\citep{Moghe2018TowardsEB}, and CMU\_DoG~\citep{Zhou2018ADF}.
The experimental results show that \textsc{GenKS} significantly improves the accuracy of knowledge selection as well as the quality of response generation, by establishing new state-of-the-art on KGD benchmarks. 
Improvements are particularly significant on unseen topics, outperforming the BART classification model by up to 8.1\% absolute.
\textsc{GenKS} also achieves the best results as the number of dialogue turns increased, with an average of 10\% improvements over the BART classification model in the last three turns.
We also compare our model with recent SOTA end-to-end methods~\citep{Shuster2021RetrievalAR}, and find our model can generate responses with fewer hallucinations while having better controllability and interpretability.
The effectiveness of the proposed method is also validated through human evaluation and ablative experiments.

Our contributions are summarized as follows:
\begin{enumerate*}[label=(\arabic*)]
    \item We propose \textsc{GenKS}, which is the first attempt at generative knowledge selection in KGD.
    \item \textsc{GenKS} captures intra-knowledge and dialogue-knowledge interactions simultaneously.
    \item We propose a hyperlink method to enhance the interactions between dialogue and knowledge.
    \item Experiments verify that \textsc{GenKS} establishes a new state-of-the-art on KGD\footnote{The code is available at: \url{https://github.com/sunnweiwei/GenKS}}.
\end{enumerate*}

%% file: sections/02-RelatedWork.tex
\section{Related work}
% The related work in this paper comprises three parts:
% (1) a brief introduction to knowledge-grounded dialogues;
% (2) the existing methods of knowledge selection;
% (3) the application of generative models to retrieval tasks.

\paragraph{Knowledge-grounded dialogues}
With the advances in large-scale language models, dialogue agents can now generate high-quality responses using parametric knowledge~\citep{Thoppilan2022LaMDALM,DeFreitas2020TowardsAH,Bao2021PLATOXLET}. 
However, hallucination remains a challenge, which means that the language model tends to generate plausible-looking statements that are factually incorrect~\citep{Shuster2021RetrievalAR}.
To address this problem, knowledge-augmented approaches are applied in dialogue generation~\citep{Lewis2020RetrievalAugmentedGF}. 
In knowledge-grounded dialogues (KGD), the dialogue models first select a knowledge snippet from passages and then generate the responses~\citep{Liu2018KnowledgeDF,Dinan2019WizardOW}.

\paragraph{Knowledge selection}
As the critical step in KGD, knowledge selection has received many studies.
The exiting methods mainly employ \emph{classification} model with dual-encoder~\citep{Dinan2019WizardOW,Kim2020SequentialLK} or cross-encoder~\citep{Zhao2020KnowledgeGroundedDG} architecture.
However, the classification paradigm is unable to capture the knowledge interaction in KGD~\citep{Kim2020SequentialLK,Li2022EnhancingKS}.
To address this problem, \citet{Li2022EnhancingKS} propose a graph-based method to capture the relationship between candidate snippets, \citet{Zhan2021CoLVAC} and \citet{Wu2021DIALKIKI} employ machine reading comprehension model to extract span from long document.
Sequential knowledge selection has also been proposed to capture the topic transition in conversations~\citep{Kim2020SequentialLK,Zhan2021AugmentingKC,Zheng2020DifferenceawareKS,Meng2020DukeNetAD,Yang2022TAKETA}.
Despite their effectiveness, the existing methods have two drawbacks: 
(1) they use compact vectors to represent dialogue and knowledge and thus lack deep semantic interactions;
(2) they are complicated and challenging to capture intra-knowledge and dialogue-knowledge interactions simultaneously.
We address these drawbacks by shifting the modeling paradigm of knowledge selection to identifier generation~\citep{Sun2022ParadigmSI}, and propose \textsc{GenKS} to capture the two types of interaction simultaneously using Transformer~\citep{Vaswani2017AttentionIA}.

\paragraph{Generative knowledge selection}
A generative paradigm for knowledge selection is not foreign to the NLP community; for example,
sequence-to-sequence models have been applied on entity retrieval~\citep{DeCao2021AutoregressiveER}, document ranking~\citep{Nogueira2020DocumentRW,Tay2022TransformerMA}, multi-evidence retrieval~\citep{Min2021JointPR,Yavuz2022ModelingMQ}, and etc.
Our proposed model \textsc{GenKS} differs from existing methods in the following ways:
(1) we are the first to explore generative knowledge selection in KGD;
(2) we consider the effectiveness of intra-knowledge interaction; 
(3) we design hyperlinks to capture the interaction between knowledge and dialogue.

%% file: sections/03-Method.tex
\section{\textsc{GenKS}}
\begin{figure*}[ht]
 \centering
 \includegraphics[width=2.0\columnwidth]{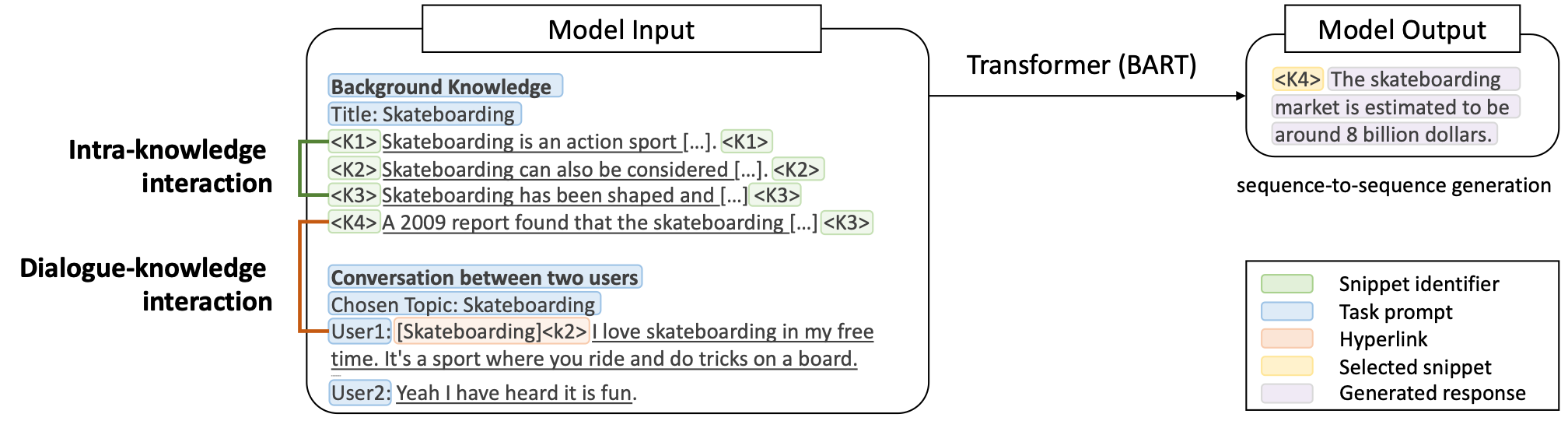}
 \caption{Overview of \textsc{GenKS}. The dialogue context and the knowledge are serialized and fed into a seq-to-seq model, BART. The outputs are the identifier of the selected snippet (i.e., \code{<k5>}) and the response.}
 \label{fig:model}
\end{figure*}

We provide an overview of \textsc{GenKS} in Figure \ref{fig:model}.
% \textsc{GenKS} can be divided into three parts: 
As shown in Figure \ref{fig:model}, the dialogue data is first serialized into a sequence. 
Then a sequence-to-sequence model (i.e., BART) is employed to select knowledge and get the response by generating the target sequence autoregressively.
In this section, we first formulate the task in Section \ref{sec:formulation}.
Then, we detail the serialization (Section \ref{sec:serialization}) and optimization (Section \ref{sec:optimization}) methods.

\subsection{Problem formulation}
\label{sec:formulation}
Suppose that we have a case of knowledge-grounded dialogues $(C,\mathcal{K},r)$, where $C=(c_1,...,c_{|C|})$ is a dialogue context that contains $|C|$ utterances, $r$ is the response to $C$,
$\mathcal{K}=(K_1,...,K_{|\mathcal{K}|})$ denotes $|\mathcal{K}|$ passages that are relevant to $C$; for each $i$, $K_i=(k_{i,1},...,k_{i,|K_i|})$ denotes a passage that contains $|K_i|$ snippets.
We define $m=\sum_{i=1}^{|\mathcal{K}|}|K_i|$ as the total number of snippets in $K$.
A knowledge-grounded dialogue agent is decoupled into two modules:
a knowledge selection module $P(k|C,\mathcal{K})$ that selects a snippet from $\mathcal{K}$;
a response generation module $P(r|C,\mathcal{K},k_{s})$ where $k_s$ is the selected snippet from knowledge selection module.

\subsection{Serialization}
\label{sec:serialization}
We formulate the knowledge selection task as a procedure of sequence generation.
As shown in Figure~\ref{fig:model}, the dialogue context $C$ and knowledge candidates $K$ are mapped into a sequence and then fed into a sequence-to-sequence model. The model's output is converted back to the selected knowledge $k$ or the response $r$.

Specifically, we first assign an identifier to each snippet in $\mathcal{K}$, sequentially starting from \code{<k1>} to \code{<km>}. 
Then we convert passages $\mathcal{K}$ into a sequence using a template that packages snippets with the corresponding identifiers and concatenates them in order; see the green block in Figure~\ref{fig:model}.
Similarly, the dialogue context $C$ is serialized by adding task prompts, i.e., task description and speaker name, as shown in the blue block in Figure \ref{fig:model}.

In multi-turn dialogues, the knowledge appearing in the dialogue history prompts the discourse structure of knowledge transition and knowledge expression.
Hence we propose a \emph{hyperlink} method to capture the dialogue-knowledge interaction explicitly.
We provide an example of the hyperlink method in Figure~\ref{fig:model}. We see that the first utterance of User1 refers to a snippet (whose identifier is \code{<k2>}) in the passage ``Skateboarding''.
We thus add a hyperlink to the utterance.
The hyperlink includes the identifier and the title of the snippet, i.e., annotating \code{[Skateboarding]<k2>} at the beginning of this utterance (as shown in the red block in Figure \ref{fig:model}).
% Note that the knowledge mentioned in previous turns is not necessarily known. 
% Generally, only the references of the knowledge used in utterances from the dialogue systems' side are explicitly known.
Finally, we splice the passages and dialogue context sequences as input for a Transformer model (i.e., BART). 
Therefore, the model can capture the intra-knowledge and dialogue-knowledge interactions through a \emph{self-attention} mechanism~\citep{Vaswani2017AttentionIA}.
% The model can decide which knowledge snippet is selected with the knowledge identifiers contained in the output sequence.

\subsection{Optimization}
\label{sec:optimization}
The knowledge selection model is optimized by the cross-entropy loss:
$
\mathcal{L}=-\log P(k_{true}|C,\mathcal{K})
$
, where $k_{true}$ denotes the label knowledge.
Since $k_{true}$ needs to be labeled manually and is not available in some scenarios~\citep{Zhou2018ADF}, we construct pseudo-labels for model training following \citet{Zhao2020KnowledgeGroundedDG} in cases the knowledge label is absent.
% Specifically, based on the intuition that human responses carry clues to the relevance of the knowledge candidates, we calculate the F1 score between each sentence and the response and utilize the knowledge with the highest score as the label.
In particular, we calculate the F1 score~\citep{Dinan2019WizardOW} between each knowledge snippet and the response.
We use the snippet with the highest score as the pseudo label.
Such a method is based on the intuition that human responses provide hints regarding the relevance of the snippets~\citep{Zhao2020KnowledgeGroundedDG,Li2020ZeroResourceKD}.

Since both knowledge selection and response generation are modeled with the \emph{generative} paradigm, 
we unify the two modules with one joint generative model.
% Regarding the response generation model, we explore an end-to-end model.
% \paragraph{End-to-end (E2E)} 
In this joint model, the knowledge selection and the response generation are optimized jointly, with shared parameters. 
To this end, we splice the knowledge identifier $k_{true}$ and response $r$ into one sequence (as shown in Figure~\ref{fig:model}).
Then, we optimize the sequence-to-sequence model using cross-entropy loss on all the tokens of the target sequence. 
In inference, the model generates knowledge identifier $k_s$ and responses $r$ in an autoregressive fashion.
We note that the end-to-end model allows the two tasks to be mutually enhanced and improves the model's efficiency.

%% file: sections/04-Experiment.tex
\section{Experimental setup}
\subsection{Datasets}
We conduct experiments on Wizard of Wikipedia (WoW)~\citep{Dinan2019WizardOW}, Holl-E~\citep{Moghe2018TowardsEB}, and CMU\_DoG~\citep{Zhou2018ADF}. 
The statistical details on these three datasets are shown in Table \ref{table:dataset} in the appendix.
\begin{itemize}[leftmargin=*,noitemsep,topsep=0pt]
\item \textbf{WoW}
is an open-domain KGD dataset using Wikipedia passage as background knowledge. 
The test set of Wizard is split into seen and unseen versions, where the unseen test set contains 58 new topics not discussed in the training data.
\item \textbf{Holl-E}
focuses on the movie domain.
The background knowledge consists of plots, comments, and movie reviews collected from different websites.
Holl-E has two versions of the test set: single test and multi-reference test. 
In the multi-reference test, there are multiple human-annotated ground-truth knowledge and corresponding responses for each instance.
% To better compare with the baselines, we adopt the modified version of Holl-E by \citet{Kim2020SequentialLK}.
%
\item \textbf{CMU\_DoG}
focuses on the domain of movies.
The workers discuss a movie in depth given the background knowledge(e.g., introduction, plots, and key scenes).
\end{itemize}
\subsection{Baselines}
We compare \textsc{GenKS} with baselines of two categories:
(i) \emph{End-to-end methods} that generate response directly without explicit knowledge selection, and 
(ii) \emph{Pipeline methods} that explicitly select knowledge snippet to be used in response.

The end-to-end methods we consider are:
\begin{itemize}[leftmargin=*,noitemsep,topsep=0pt]
\item \textbf{BART}~\citep{Lewis2020BARTDS} that generates responses without access to the external passage and uses knowledge inside model parameters instead.
\item \textbf{BART~FID}~\citep{Izacard2021LeveragingPR} concatenates and encodes each candidate knowledge with dialogue separately and fuses all the encoded representation in the decoder to generate the response.
\item \textbf{BART~RAG-DPR} is a baseline adopted by \citet{Adolphs2021ReasonFT}, which uses DPR-retrieved passages and produces response usning RAG.
\item \textbf{BART~FiD-RAG~DPR-Poly}~\citep{Shuster2021RetrievalAR} uses DPR-Poly to retrieve passage and uses FiD-RAG to generate the response.
\end{itemize}

\noindent Regarding the pipeline baselines, according to their knowledge selection modeling paradigm, we sub-categorize pipeline baselines into four groups:\\
\noindent(1) The \emph{Classification methods} includes: 
\begin{itemize}[leftmargin=*,noitemsep,topsep=0pt]
\item \textbf{SKT}~\citep{Kim2020SequentialLK} proposes sequential knowledge selection. 
\item \textbf{DiffKS}~\citep{Zheng2020DifferenceawareKS} captures the knowledge differences between adjacent turns. 
\item \textbf{DukeNet}~\citep{Meng2020DukeNetAD} models the knowledge shift and tracking processes with a dual learning scheme. 
\item \textbf{KnowledGPT}~\citep{Zhao2020KnowledgeGroundedDG} exploits pre-trained language models in KGD. 
\item \textbf{MIKe}~\citep{Meng2021InitiativeAwareSL}  distinguish user-initiative and system-initiative. 
\item \textbf{K-Mine}~\citep{Lotfi2021TeachMW} proposes a score-and-aggregate module.
\item \textbf{TAKE}~\citep{Yang2022TAKETA} propose a topic-shift aware network.
\end{itemize}
(2) The \emph{MRC methods} includes: 
\begin{itemize}[leftmargin=*,noitemsep,topsep=0pt]
\item \textbf{CoLV}~\citep{Zhan2021CoLVAC} proposes a collaborative latent variable model.
\item \textbf{DIALKI}~\citep{Wu2021DIALKIKI} proposes a MRC-based model to extract span from passage.
\end{itemize}
(3) The \emph{Graph-based methods} includes:
\begin{itemize}[leftmargin=*,noitemsep,topsep=0pt]
\item \textbf{Graph}~\citep{Li2022EnhancingKS} builds a semantic graph upon candidate documents and employs a GNN model. And
\end{itemize}
(4) The \emph{Knowledge generation methods} includes: 
\begin{itemize}[leftmargin=*,noitemsep,topsep=0pt]
\item \textbf{K2R}~\citep{Adolphs2021ReasonFT} uses the RAG-based model to generate knowledge text and then generates dialogue response based on it. 
\end{itemize}

\subsection{Evaluation metrics}
% We choose the automatic evaluation metrics following the conventions of the dataset.
In WoW, we choose perplexity (\textbf{PPL}) of the ground-truth responses, unigram \textbf{F1}\footnote{\url{https://github.com/facebookresearch/ParlAI}}~\citep{Dinan2019WizardOW}, \textbf{Knowledge-F1}~\citep{Shuster2021RetrievalAR}, and {\textbf{BLEU-4}}~\citep{Papineni2002BleuAM} score as metrics.
In Holl-E, we additionally use \textbf{ROUGE-1}, and \textbf{ROUGE-2} following \citet{Meng2020DukeNetAD}.
In CMU\_DoG, we additionally use embedding-based metrics includes \textbf{Average}, \textbf{Extreme}, and \textbf{Greedy} following \citet{Zhao2020KnowledgeGroundedDG}.

In addition, we randomly sample 100 examples from the WoW test seen and WoW test unseen, respectively, and recruit three experts for human evaluation. 
The annotators are asked to judge the model-generated response in four ways:
\begin{itemize}[leftmargin=*,noitemsep,topsep=0pt]
    \item \textbf{Fluency}, which measures whether the response is fluency in expression; 
    \item \textbf{Coherence}, which measures whether the response is coherence to the dialogue context; 
    \item \textbf{Relevance}, which measures whether
    the knowledge used in the response is relevant to the dialogue; 
    and 
    \item \textbf{Factuality} measures whether the response's content is factual.
    In Factuality evaluation, the experts check the content using Google.
\end{itemize}
The annotators are asked to assign a score in \{0, 1\} (representing ``nonfactual'' and ``factual'') for factuality, and a score in \{0, 1, 2\}
(representing ``bad'', ``fair'', and ``good'') for the others.

\subsection{Implementation details}
We implement the \textsc{GenKS} using BART large (with 400M parameters)~\citep{Lewis2020BARTDS} in HuggingFace's Transformers library. 
We truncate the dialogue context to 256 tokens, then truncate the knowledge so that the total length is less than 1024 tokens.
During inference, the responses are decoded using a greedy search. See Appendix~\ref{appendix:implementation} for more details.

Typically, the number of passages in $\mathcal{K}$ is large, so that the input sequence exceeds the maximum input length of BART (i.e., 1024 tokens).
To address this problem, we take advantage of a lightweight passage selector based on DistilBERT (with 66M parameters)~\citep{Sanh2019DistilBERTAD}, which aims to rank the passages in $\mathcal{K}$.
Specifically, we concatenate each passage with dialogue context and encode the sequence using DistilBERT. 
Finally, the representation of \code{[CLS]} token is used to estimate the relevance score of the passage through a learnable MLP classifier.
The passage selector is optimized via contrastive learning objective~\citep{Nogueira2019PassageRW}, in which the model learns to assign a higher score to positive passages than negative passages.
During inference, we keep only the top-1 passage ranked by the passage selector.
The passage selector gets Recall@1 of 75.5\%, 76.5\%, and 68.0\% for the WoW test seen, WoW test unseen, and Holl-E, respectively.

%% file: sections/05-Results.tex
\input{tables/ks}

\input{tables/gen-wow}

\section{Experimental results}

\subsection{Performance on knowledge selection}
We evaluate the knowledge selection effectiveness of \textsc{GenKS} on WoW and Holl-E, respectively\footnote{We are unable to evaluate the knowledge selection accuracy on CMU\_DoG accuracy of knowledge selection on CMU\_DoG due to the lack of manual labeling.}.
In Table \ref{table:ks}, we compare the knowledge selection accuracy of \textsc{GenKS} with previous pipeline methods. 
Results show that \textsc{GenKS} outperforms the baselines and achieves the highest accuracy of knowledge selection on both datasets.

We find that \textsc{GenKS} particularly excels at topics that do not appear in training dataset, as evidenced by its performance on the WoW unseen test split. 
In comparison, the classification models show a noticeable decrease in accuracy on the unseen topic. 
We also find that the baselines that incorporate intra-knowledge interaction, such as, \textsc{GenKS}, Graph, DIALKI, have a better grasp of knowledge of unseen topics\footnote{Note that the higher accuracy results on unseen than seen might be due to the smaller number of topics included in the unseen test set.}.

\begin{figure}[t]
 \centering
 \includegraphics[width=1\columnwidth]{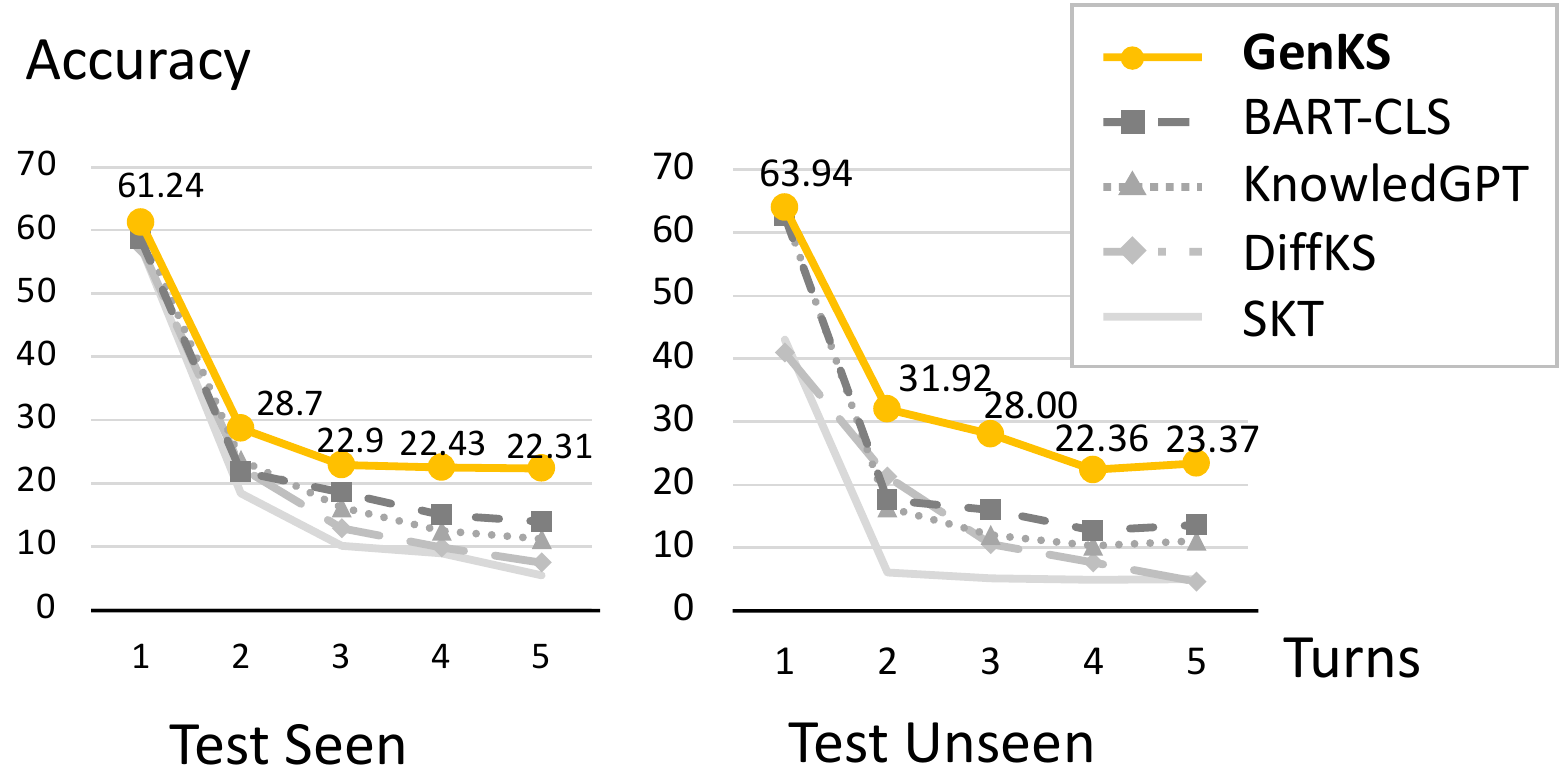}
 \caption{Knowledge selection accuracy over different dialogue turns. 
  BART-CLS represents a text-matching model with cross-encoder architecture.}
 \label{fig:turn}
\end{figure}

To evaluate the performance of \textsc{GenKS} as dialogue goes deeper, we compare \textsc{GenKS} with four classification baselines (SKT, DiffKS, KnowledGPT, and BART-CLS) overturns in Figure \ref{fig:turn}. 
We find that in the early stages of the conversation, both \textsc{GenKS} and the baseline methods achieve high accuracy.
However, as the dialogue dives deeply into a topic, a significant performance decline is observed among the baseline methods.
In contrast, \textsc{GenKS}, which explicitly captures the interaction between multi-turn dialogue and knowledge, maintained a relatively high accuracy (around 22\%-23\%).

\subsection{Quality of generated responses}
% \paragraph{Comparison with previous methods}
We report response generation evaluation results on WoW in Table~\ref{table:gen-wow}. The results on Holl-E and CMU\_DoG are available in Table~\ref{table:gen-holle} and Table~\ref{table:gen-cmu} in the appendix. 
The results of baselines are cited from original papers or re-evaluated using officially released checkpoints. 
% We use WoW as the main dataset for analysis because most baselines use WoW as the benchmark data.

Compared with previous pipeline models, \textsc{GenKS} achieve the best performance on almost all metrics.
For example, \textsc{GenKS} surpasses KnowledGPT by 0.7\% and 2.4\% in terms of F1 on WoW seen and WoW unseen, respectively.
% \textsc{GenKS} is 
Note that the improvements on the unseen test set are more notable than on the seen test set, which agrees with the experimental results regarding knowledge selection.
\textsc{GenKS} also achieve competitive results compared to SOTA end-to-end models.
For example, \textsc{GenKS} performs comparably to BART~FiD-RAG~DPR-Poly on WoW seen and outperformed on WoW unseen.

\subsection{Ablation study about knowledge selection}
\label{sec:ablation}
To analyze the effect of each component in \textsc{GenKS}, we designed several variants and conducted an ablation study about knowledge selection.
Results are listed in Table~\ref{table:ks}, ``Variants for comparison''.
The details of compared variants and the findings are as follows:

\noindent \textbf{BART classification}
We use BART to classify each candidate snippet into two classes: ``relevant'' or ``irrelevant''.
The results show that BART in the classification paradigm performs worse than \textsc{GenKS} by a large margin.

\noindent \textbf{BART classification w/ position}
To understand the influence of position bias,
we splice the snippet's position into the classification model's input.
We find that the results are improved to a certain extent (about 1\% improvement), but there is still a clear gap compared with \textsc{GenKS}.

\noindent \textbf{Hierarchical classification}
This variant first uses the passage selector model of \textsc{GenKS} to rank the passages and then selects the snippets in the top-ranked passage using BART classification w/ position.
The results show that the passage selector does not affect the classification model's performance.

\noindent \textbf{Without passage selector}
When the passage selector model of \textsc{GenKS} is removed, the model has more probability of truncating the label knowledge, resulting in an evident decline in performance.

\noindent \textbf{Unorder knowledge snippets} 
To disable the intra-knowledge interaction, we unorder the snippets so that order of the snippets is inconsistent with the original passages.
% We shuffle the snippets in the passages so that the structural information of the passage is corrupted. 
This variant shows a decline in selection accuracy, especially on unseen topics, indicating that keeping the order of the snippets in the passage is necessary.

\noindent \textbf{Without hyperlinks} 
We remove the hyperlinks in the dialogue context. 
About a 1\% accuracy drop is seen, indicating the effectiveness of hyperlinks.

\subsection{Ablation study about response generation}
\label{sec:ablation-gen}
As shown in Table \ref{table:gen-wow}, we also conduct an ablation study about response generation.
The details of compared variants and the findings are as follows:

\noindent \textbf{With BART classification knowledge} 
When replacing the generated identifier with the knowledge selected by BART classification,  a performance decline is witnessed --the F1 value drops by 0.7\% and 1.8\% on Wizard seen and unseen, sustaining the effectiveness of the knowledge selection of \textsc{GenKS}.

\noindent \textbf{Without identifier generation} 
This variant removes the identifier generation by directly generating the response.
We see notable performance drops, especially in the KF1 metric.
The results indicate that explicit training and inference about knowledge selection enable to use of more appropriate knowledge in response generation.

\noindent \textbf{Without hyperlinks} 
This variant removes hyperlinks from \textsc{GenKS}.
It performs worse than \textsc{GenKS}, probably due to its lower accuracy of knowledge selection than \textsc{GenKS}.

\noindent \textbf{Use the oracle knowledge} 
We replace the model-predicted snippet identifier with the oracle one (knowledge used by ground-truth response). 
The results (e.g., KF1=74) suggest that \textsc{GenKS} can effectively locate and incorporate the corresponding knowledge into the responses following the guidance of the identifier.

\input{tables/human}

\subsection{Human evaluation}
Table \ref{table:human} shows the human-evaluating results. 
Results show that \textsc{GenKS} consistently outperforms baselines on all datasets. 
The Fleiss' kappa value is above $0.60$, indicating substantial agreement among the annotators.
\textsc{GenKS} outperforms KnowledGPT by about 0.02 and DukeNet by about 0.20 in terms of response generation evaluation metrics (i.e., \emph{Fluency} and \emph{Context Coherence}).
Moreover, for the \emph{Knowledge Relevance}, the annotators agree that \textsc{GenKS} is capable of selecting knowledge that is more relevant to the dialogue and generating more informative responses than baselines.
The \emph{Factuality} results show that by explicitly identifying the knowledge snippet used in response, \textsc{GenKS} can reduce the hallucination of response generation.

\begin{figure}[t]
 \centering
 \includegraphics[width=1\columnwidth]{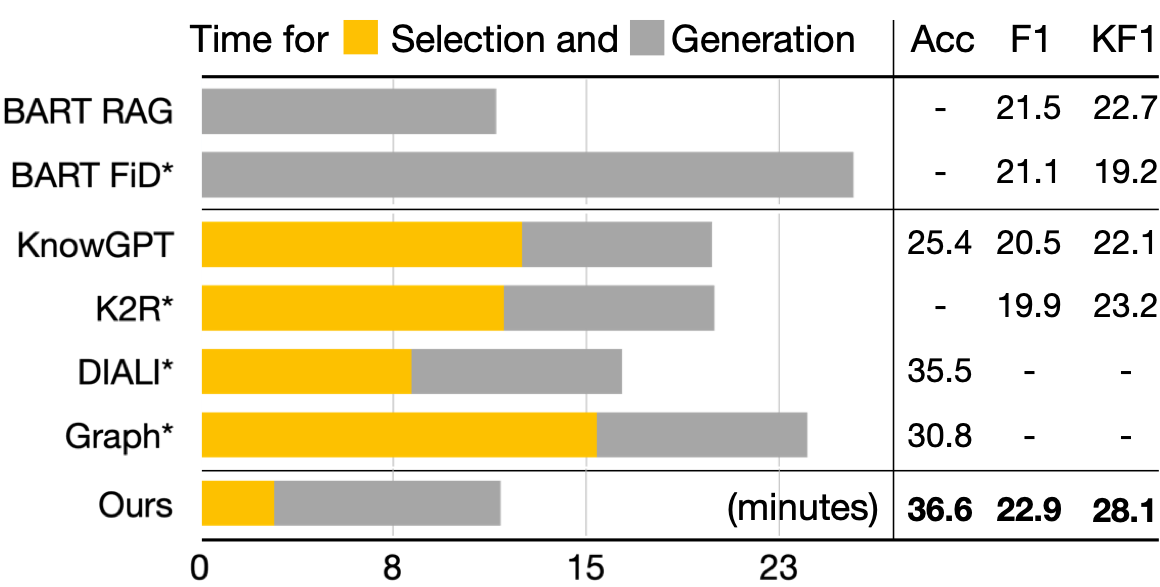}
 \caption{Inference time (minutes) on one GPU on WoW unseen test set. The values of models with $^*$ are estimated based on the model size and input/output length.
 }
\label{fig:speed}
\end{figure}

\subsection{Efficiency evaluation}
To evaluate the efficiency of \textsc{GenKS}, we compare it with previous end-to-end models and pipeline models.
The results are presented in Figure~\ref{fig:speed}, which shows that \textsc{GenKS} is more efficient than previous pipeline models.
This may be because \textsc{GenKS} integrates knowledge selection and response generation in an end-to-end model, avoiding repeated encoding of dialogue history and background knowledge.
Furthermore, we find that \textsc{GenKS}, as a pipeline method, achieves comparable efficiency to end-to-end models like RAG, but with the added advantage of explicit knowledge selection.

\input{tables/multi.tex}
\subsection{Analytical experiment}

\paragraph{Multi-snippets selection}
\textsc{GenKS} initially selects a single snippet following the experimental setup outlined in the baselines~\citep{Dinan2019WizardOW}, but can also select multiple snippets by generating multiple identifiers.
To evaluate the performance of this approach, we test a variant of the \textsc{GenKS} model, namely \textsc{GenKS-2}, which selects two snippets by generating two identifiers consecutively. 
We compare its performance with the original \textsc{GenKS} on the WoW dataset.
The results are listed in Table~\ref{table:multi} group 1. 
\textsc{GenKS-2} performs slightly worse than the original \textsc{GenKS}, likely because the WoW dataset only uses one snippet in response annotation and therefore does not benefit from using multiple snippets~\citep{Dinan2019WizardOW}. 
Nevertheless, the results suggest that the proposed generative knowledge selection approach has the ability to select multiple knowledge.

\paragraph{Hyper-parameter analysis}
We also conduct ablation experiments on the number of input snippets to the model and maximum input tokens.
The results are listed in Table~\ref{table:multi} group 2.
We find that reducing the number or length of knowledge reduces model effectiveness.

\begin{table}[!t]
\small \centering
\setlength\tabcolsep{2pt}
\begin{tabular}{@{}rp{6.5cm}@{}}

\toprule
\textbf{Topic} & Budweiser\\
\textbf{User} & I think Budweiser taste terrible. Have you ever had it?\\
\midrule
\textbf{Know 1} &{Produced in various breweries around the world, Budweiser is a filtered beer available in draft and packaged forms.}\\
\textbf{Res 1} &{Yes, I have.  It is produced in various breweries around the world!}\\
\midrule
\textbf{Know 2} & Budweiser is an American-style pale lager produced by Anheuser-Busch, currently part of the transnational corporation Anheuser-Busch InBev.\\
\textbf{Res 2} & Yes, I have. I know that it is an American-style pale lager produced by Anheuser-Busch.\\
\bottomrule
\end{tabular}
\caption{Examples of \textsc{GenKS} outputs on the WoW.} \label{table:case0}
\end{table}

\subsection{Case study}
To better understand end-to-end baselines and our model, we provide an example in Table~\ref{table:case0}, which shows that \textsc{GenKS} appropriately changes its response prediction when providing different knowledge snippets\footnote{Note that this example only aims to show the output of the model. In fact, according to \url{https://en.wikipedia.org/wiki/Budweiser}, Budweiser is also a famous lager from the Czech Republic, and American Budweiser is sold and known as Bud through most of the European Union.}.
Therefore, \textsc{GenKS} is more controllable and interpretable than end-to-end models, where the end-to-end system is a black box.
We provide more case studies in Appendix~\ref{appendix:case}.

%% file: tables/ks.tex
\begin{table}
\centering
\small
\setlength\tabcolsep{1pt}
\begin{tabular}{@{}l cccc@{}}

\toprule
\multirow{2}{*}{\textbf{Methods}}
% {\textbf{Methods}}
& \multicolumn{2}{c}{\textbf{WoW}} 
& \multicolumn{2}{c}{\textbf{Holl-E}}
\\

\cmidrule(lr){2-3} \cmidrule(lr){4-5}

{} & \textbf{Seen} & \textbf{Unseen} & \textbf{Single} & \textbf{Multi}\\
\midrule
\emph{Classification methods}\\
% TMN~\citep{Dinan2019WizardOW} & 22.5 & 12.2 & 22.7 & 32.3\\
SKT~\citep{Kim2020SequentialLK} & 26.8 & 18.3 & 29.2 & 39.2\\
DukeNet~\citep{Meng2020DukeNetAD} & 26.4 & 19.6 & 30.0 & 40.3\\
DiffKS~\citep{Zheng2020DifferenceawareKS} & 25.5 & 19.7 & 33.0 & - \\
KnowledGPT~\citep{Zhao2020KnowledgeGroundedDG} & 28.0 & 25.4 & - & -\\
MIKe~\citep{Meng2021InitiativeAwareSL} & 28.4 & 21.5 & 31.9 & 41.8\\
K-Mine~\citep{Lotfi2021TeachMW} & 29.7 & 28.3 & 31.7 & -\\
TAKE~\citep{Yang2022TAKETA} & 28.8 & 25.8 & - & - \\
% RoBERTa & 29.5 & 29.1 & 34.5 & 44.3\\
\midrule
\emph{Other methods}\\
CoLV~\citep{Zhan2021CoLVAC} & 30.1 & 18.9 & 32.7 & -\\
DIALKI~\citep{Wu2021DIALKIKI} & \underline{32.9} & \underline{35.5} & - & - \\
Graph~\citep{Li2022EnhancingKS} & 29.4 & 30.8 & 
\underline{37.7} & \underline{46.1} \\
\midrule
\textbf{GenKS} & \textbf{34.2} & \textbf{36.6} & \textbf{37.9} & \textbf{46.8} \\
% \multicolumn{11}{c}{\emph{retrieval-free approaches}}\\
% \hline
% DialoGPT$_{ft}$~\citep{Zhao2020ArePL} & 16.2 & 19.0 & 0.871 & 0.461 & 0.683 & 20.4 & 17.6 & 0.869 & 0.451 & 0.674 \\
% KnowExpert~\citep{Xu2021RetrievalFreeKD} & 15.3 & 18.7 & - & - & - & 20.1 & 16.7 & - & - & - \\
% Megatron~\citep{Liu2022MultiStagePF} & - & 19.5 & 0.840 & 0.427 & 0.726 & - & 16.8 & 0.831 & 0.429 & 0.723 \\
% BlenderBot~\citep{Roller2021RecipesFB} & 8.7 & 18.8 & - & - & - & 10.4 & 17.8 & - & - & - \\
% BART~\citep{Lewis2020RetrievalAugmentedGF} & 14.7 &  20.9 & - & - & - & 18.9 &  18.7 & - & - & - \\
\midrule
\emph{Variants for comparison}\\
{-~BART classification} & 29.8 & 29.7 & 34.0 & 44.0\\
{-~BART classification w/ position} & 30.1 & 31.2 & 34.0 & 44.0\\
{-~Hierarchical classification} & 30.0 & 31.4 & 33.8 & 43.7\\
{-~Without passage selector} & 31.4 & 32.0 & 34.5 & 44.4\\
{-~Unorder knowledge snippets} & 31.8 & 33.3 & 36.5 & 45.8\\
{-~Without hyperlink} & 33.4 & 35.4 & 36.9 & 45.4\\
% {-~Without semantic enhanced} & 33.1 & 35.9 & 36.7 & 45.1\\

\bottomrule
\end{tabular}
\caption{Knowledge selection accuracy on WoW (seen and unseen test set) and Holl-E (single reference and multi-reference test set). 
\textbf{Bold} denote the best results with significant improvements over the previous SOTA (t-test, $p<0.05$). \underline{Underline} denote second best results.}
\label{table:ks}
\end{table}

%% file: tables/gen-wow.tex
\begin{table*}[!t]
\centering
\small \setlength\tabcolsep{5pt}
\begin{tabular}{@{}l cccc cccc@{}}

\toprule
% \hline
\multirow{2}{*}{\textbf{Methods}}
% {\textbf{Methods}}
& \multicolumn{4}{c}{\textbf{WoW Seen}} 
& \multicolumn{4}{c}{\textbf{WoW Unseen}}
\\
% \cline{2-11}
\cmidrule(lr){2-5} \cmidrule(lr){6-9}

{} & PPL & F1 & KF1 & B4 & PPL & F1 & KF1 & B4 \\

\midrule
\emph{End-to-end models}\\
BART~\citep{Lewis2020BARTDS}
& 14.7 & 20.9 & 17.4 & 1.7
& 18.9 & 18.8 & 15.1 & 0.9
\\
% BART Passage
% & - & 22.1 & 25.2 & 3.7
% & - & 22.3 & 25.3 & 4.1
% \\
BART FiD~\citep{Izacard2021LeveragingPR}
& 17.0 & 21.5 & 20.0 & 3.6
& 18.4 & 20.6 & 19.2 & 3.2
\\
BART RAG-DPR~\citep{Adolphs2021ReasonFT}
& \textbf{11.5} & 22.6 & 26.1 & 3.7 
& \textbf{13.1} & 21.5 & 22.7 & 3.0
\\
BART FiD-RAG DPR-Poly~\citep{Shuster2021RetrievalAR}
& \textbf{11.4} & 22.1 & \textbf{29.7} & \textbf{4.1}
& \textbf{13.1} & 21.1 & 27.1 & 3.8
\\
\midrule
\emph{Pipeline models}\\
% TMN~\citep{Dinan2019WizardOW}
% & 66.5 & 16.9 & 15.8 & 0.4
% & 97.3 & 14.3 & 9.4 & 0.3
% \\
DukeNet~\citep{Meng2020DukeNetAD}
& 48.3 & 19.3 & 18.5 & 2.4
& 69.4 & 17.1 & 16.5 & 1.7 
\\
CoLV~\citep{Zhan2021CoLVAC}
& 39.5  & 20.3 & 18.2 & 2.8
& 54.3 & 18.5 & 17.5 & 2.1
\\
KnowledGPT~\citep{Zhao2020KnowledgeGroundedDG}
& 19.2  & 22.0 & 23.8 & 3.7
& 22.3 & 20.5 & 22.1 & 3.0
\\
K-Mine~\citep{Lotfi2021TeachMW}
& 13.2 & 21.8 & - & - 
& 16.4 & 21.1 & - & -
\\
K2R RAG-DPR~\citep{Adolphs2021ReasonFT} 
& 18.3 & 22.0 & 27.3 & 3.7 
& 22.3 & 19.9 & 23.2 & 2.8
\\
K2R BART RAG-DPR~\citep{Adolphs2021ReasonFT} 
& 17.9 & 21.3 & \textbf{29.2} & 3.5
& 21.1 & 19.9 & 24.3 & 2.5
\\
% UniGDD~\citep{Gao2022UniGDDAU}
% & 16.5 & 21.2 & 26.2 & 3.6 
% & 18.4 & 20.6 & 25.5 & 3.2
% \\
\midrule

\textbf{GenKS} 
& 13.1 & \textbf{22.9}\rlap{$^{*}$} & \textbf{29.5} & \textbf{4.5}\rlap{$^{*}$}
& \textbf{13.2} & \textbf{22.7}\rlap{$^{*}$} & \textbf{28.1}\rlap{$^{*}$} & \textbf{4.6}\rlap{$^{*}$}
\\
\midrule
\emph{Ablative variants}\\
% - Remove historical identifiers
% & 
% \\
- With BART classification knowledge
& 14.7 & 22.0 & 25.9 & 3.5
& 16.2 & 21.1 & 24.4 & 3.1
\\
% - Use separate model 
% & 14.6 & 22.3 & 
% & 14.7 & 22.4 & 
% \\
% - Remove identifier generation
% & 16.8 & 21.7 & 25.2 & 3.7 
% & 17.3 & 21.8 & 23.3 & 3.9
% \\
- Without identifiers generation
& 13.8  & 21.7 & 23.2 & 3.7 
& 14.1  & 21.8 & 23.3 & 3.9
\\
- Without hyperlink
& 14.2 & 22.1 & 27.2 & 3.9 
& 15.5 & 22.3 & 26.9 & 4.2
\\
-~With oracle knowledge
& 8.9 & 38.8 & 74.2 & 13.1
& 10.5 & 38.9 & 74.5 & 12.8
\\
\bottomrule
\end{tabular}
\caption{Evaluation results on WoW seen and unseen test set in terms of response quality. 
We compare against the ground-truth dialogue response in terms of perplexity (PPL), F1, Knowledge F1 (KF1), and BLEU-4 (B4).
The four groups lists previous end-to-end models, previous pipeline models, GenKS, and ablative variants.
The best results are highlighted with \textbf{bold}, and the second-best results are highlighted with \underline{underline}. * indicates significant improvements over all baselines with p-value $< 0.05$.}
\label{table:gen-wow}
\end{table*}

%% file: tables/human.tex
\begin{table}[!t]
\centering
\small
\setlength\tabcolsep{2pt}
\begin{tabular}{@{}l cccc cccc@{}}

\toprule
% \hline
\multirow{2}{*}{\textbf{Methods}}
% {\textbf{Methods}}
& \multicolumn{4}{c}{\textbf{WoW Seen}} 
& \multicolumn{4}{c}{\textbf{WoW Unseen}}
% & \multicolumn{3}{c}{\textbf{Holl-E}}
\\
% \cline{2-11}
\cmidrule(lr){2-5} \cmidrule(lr){6-9}

{} & Flu. & Coh. & Rel. & Fact. & Flu. & Coh. & Rel. & Fact. \\

\midrule
BART
& 1.82 & 1.51 & 1.45 & 0.82
& 1.76 & 1.50 & 1.47 & 0.76
\\
BART FiD
& 1.88 & 1.70 & 1.55 & 0.84
& 1.85 & 1.67 & 1.53 & 0.82
\\

\midrule
TMN
& 1.59 & 1.41 & 1.08 & 0.62
& 1.42 & 1.30 & 0.98 & 0.59
% & 1.77 & 1.57 & 1.15
\\

DukeNet
& 1.69 & 1.56 & 1.22 & 0.71
& 1.66 & 1.47 & 1.10 & 0.72
% & 1.81 & 1.61 & 1.51
\\

KnowledGPT
& 1.89 & 1.67 & 1.58 & 0.87
& 1.87 & 1.68 & 1.51 & 0.83
% & - & - & -
\\

\textbf{GenKS} 
& \textbf{1.90} & \textbf{1.72} & \textbf{1.69} & \textbf{0.89}
& \textbf{1.91} & \textbf{1.71} & \textbf{1.67} & \textbf{0.91}
% & \textbf{1.90} & \textbf{1.75} & \textbf{1.72}
\\

\bottomrule
\end{tabular}
\caption{Human evaluation results. Flu, Coh, Rel, and Fact denote Fluency, Coherence, Relevance, and Factuality, respectively.}
\label{table:human}
\end{table}

%% file: tables/multi.tex
\begin{table}[!t]
\centering
\small
\setlength\tabcolsep{4.5pt}
\begin{tabular}{@{}l ccc ccc@{}}

\toprule
% \hline
\multirow{2}{*}{\textbf{Methods}}
% {\textbf{Methods}}
& \multicolumn{3}{c}{\textbf{WoW Seen}} 
& \multicolumn{3}{c}{\textbf{WoW Unseen}}
\\
% \cline{2-11}
\cmidrule(lr){2-4} \cmidrule(lr){5-7}

{} & F1 & KF1 & B4 & F1 & KF1 & B4 
\\

\midrule
GenKS
& 22.9 & 29.5 & 4.5 & 22.7 & 28.1 & 4.6
\\
GenKS-2
& 22.4 & 29.3 & 4.2 & 22.2 & 27.6 & 4.2
\\

\midrule

GenKS (5 Snippets) & 22.3 & 27.6 & 4.2 & 21.8 & 25.5 & 4.1\\
GenKS (3 Snippets)  & 21.1 & 29.3 & 3.2 & 20.0 & 20.9 & 2.9 \\
GenKS (128 Tokens) & 21.5 & 25.6 & 3.5 & 20.7 & 22.9 & 3.4 \\
GenKS (64 Tokens)  & 20.7 & 23.3 & 3.0 & 20.1 & 20.6 & 2.9 \\

\bottomrule
\end{tabular}
\caption{Analytical experiment results on WoW. The first group compares \textsc{GenKS} and its variant \textsc{GenKS}, which selects two snippets instead of one. The second group includes the results of \textsc{GenKS} with different maximum number of snippets inputs or maximum input tokens.}
\label{table:multi}
\end{table}

%% file: sections/06-Conclusion.tex
\section{Conclusion}
In this paper, we have proposed \textsc{GenKS}, a simple yet effective knowledge-grounded dialogue model.
\textsc{GenKS} is a generative model, which learns to select knowledge snippets by generating their identifiers.
Benefiting from the modeling of intra-knowledge interaction and dialogue-knowledge interaction, \textsc{GenKS} effectively addresses the challenges of \emph{ambiguity} and \emph{discourse structure}.
% \textsc{GenKS} . 
Our experiments have shown that \textsc{GenKS} establishes a new state-of-the-art on three knowledge-grounded dialogue benchmarks.
Notably, \textsc{GenKS} particularly excels at new topics and as the dialogue goes deeper.
\textsc{GenKS} also outperforms SOTA end-to-end models.
Hence, we believe \textsc{GenKS} reveals a new paradigm for knowledge selection in open-domain dialogue.

\section*{Limitations}
The limitations of this work include the modular modeling of passage reranks, which reduces the efficiency of the approach.
Besides, we only conduct human evaluation on one popular dataset, i.e., Wizard of Wikipedia.
Furthermore, the effectiveness of \textsc{GenKS} is only verified in the English dataset. 
Research on other languages establishes a new challenge, especially for languages with limited knowledge and annotated data.
In future work, we would like to explore more efficient passage rerank techniques on knowledge-grounded dialogues.
We will also conduct human evaluation for more datasets.
Besides, generative knowledge selection can be extended to future studies about conversational recommendation.

\section*{Ethics statement}
The paper proposes a knowledge-grounded dialogue system to generate a response using external knowledge.
The intended use of this system is to perform chit-chat with the user on topics such as books and movies. 
The system is developed using large pre-trained language models (i.e., BART), who are trained on large-scale web data known to contain biased or discriminatory content.
The datasets (i.e., WoW, Holl-E, CMU\_DoG) that we train on also include subjective knowledge (comments on movies) that may express the bias of the writers.
Although the system is able to reduce the hallucination of response compared to end-to-end models, the outputs from our system may still contain non-factual information and should not be considered as advice for any critical decision-making.

\section*{Acknowledgements}
This work was supported by the National Key R\&D Program of China with grant No. 2020YFB1406704, the Natural Science Foundation of China (62272274, 62202271, 61902219, 61972234, 62072279, 62102234), the Natural Science Foundation of Shandong Province (ZR2021QF129), the Key Scientific and Technological Innovation Program of Shandong Province (2019JZZY010129).
All content represents the opinion of the authors, which
is not necessarily shared or endorsed by their respective
employers and/or sponsors.

%% file: sections/07-Appendix.tex
% \begin{appendices}
\appendix

\section{Implementation details}
\label{appendix:implementation}
We use gradient clipping with a maximum gradient norm of $0.1$. 
We optimize the model for up to 5 epochs with a batch size of 16 on 4 3090 GPUs with 24G memory.
We choose the model checkpoints by evaluating the metrics on the validation set for each epoch.
During inference, the responses are decoded using a greedy search. We have tried some advanced decoding algorithms (e.g., nucleus sampling) and found no improvement.
The training of the model can be completed within 5h, and the latency of the model inference for one example is within 0.1s.
The passage rerank model gets Recall@1 of 75.5\%, 76.5\%., 61.0\% for WoW test seen, WoW test unseen, and Holl-E, respectively.

\input{tables/dataset}

\input{tables/gen-holle}
\input{tables/gen-cmu}

\section{Case study}
\label{appendix:case}
To better understand baselines and our model, we present two examples in Table~\ref{table:case1} and Table~\ref{table:case2}.
Table~\ref{table:case1} show example where both \textsc{GenKS} and baselines select out the proper knowledge (i.e., the knowledge snippet shown in green).
We see that the response generated by \textsc{GenKS} is more appropriate to the dialogue context than baselines, while KnowledGPT's response does not answer User2's question and is also factually incorrect. 
In Table~\ref{table:case2}, we observed that although neither \textsc{GenKS} nor the baselines selected the label knowledge, the response generated by \textsc{GenKS} is still more natural and coherence.
We also find that KnowledGPT is more colloquial than \textsc{GenKS} but has problems with hallucinations. 

% \begin{figure}[t]
%  \centering
%  \includegraphics[width=1\columnwidth]{figs/case.pdf}
%  \caption{Case study on the Wizard Test Unseen dataset. (a) shows an example where both \textsc{GenKS} and baselines select the proper knowledge (b) shows an example where both models select the wrong knowledge.}
%  \label{fig:case}
% \end{figure}

\input{tables/case}
% \end{appendices}

%% file: tables/dataset.tex
\begin{table}[!t]
\centering
\small
\setlength\tabcolsep{5pt}
\begin{tabular}{@{}l ccc ccc@{}}

\toprule
& \textbf{WoW} & \textbf{Holl-E} & \textbf{CUM\_DoG}  \\
\midrule
% \cline{2-11}

Training size & 18,430 & 7,228 & 3,373 \\
Validation size  & 1,948 & 930 & 229\\
Test size & 965 / 968 & 913 & 619\\
Number of topics & 1,365 & 858 & 30\\
Avg. Turn per dialogue & 9.0 & 10.1 & 22.2\\
Avg. Num of snippets & 62.5 & 57.3 & 36.3\\
Avg. Num of passages & 11.6 & 5.9 & 4.0\\
\bottomrule
\end{tabular}
\caption{Statistics of two experimental datasets, Wizard
of Wikipedia (WoW), Holl-E, and CMU\_DoG. 
The two numbers in WoW indicate the size of seen and unseen test set, respectively.}
\label{table:dataset}
\end{table}

%% file: tables/gen-holle.tex
\begin{table}[!t]
\centering
\small
\setlength\tabcolsep{3pt}
\begin{tabular}{@{}l ccccc@{}}

\toprule
{} & F1 & B4 & KF1 & RG1 & RG2 \\

\midrule
BART~\citep{Lewis2020BARTDS}
& 34.7 & 22.3 & 29.1 & 38.0 & 27.9
\\
% BART Passage
% & 34.7 & 22.3 & 38.0 & 27.9
% \\
\midrule

CoLV~\citep{Zhan2021CoLVAC}
& - & 20.3 & - & 32.0 & 25.8
\\

MIKe~\citep{Meng2021InitiativeAwareSL}
& 32.1 & 21.1 & - &  38.0 & 25.2
% & 38.3 & 28.1 & 44.1 & 31.5 
\\
Graph~\citep{Li2022EnhancingKS}
& - & - & - & \textbf{42.5} & 34.4
% & 38.3 & 28.1 & 44.1 & 31.5 
\\
\midrule

% \textbf{GenKS 2SG} 
% & 32.7 & \textbf{22.3} & \textbf{38.0} & 27.9
% \\

\textbf{GenKS} 
& \textbf{36.7} & \textbf{24.3} & \textbf{31.3} & \textbf{42.3} & \textbf{35.2}
% & \textbf{39.5} & \textbf{29.7} & \textbf{44.9} & \textbf{34.4}
% & 14.6 & 22.3 & 4.3
\\

\bottomrule
\end{tabular}
\caption{Results on Holl-E in term of response quality. RG1 and RG2 denote ROUGE-1 and ROUGE-2 respectively.
Best results are heighten with \textbf{bold}.}
\label{table:gen-holle}
\end{table}

%% file: tables/gen-cmu.tex
\begin{table}[ht]
\centering
\small
\setlength\tabcolsep{2pt}
\begin{tabular}{@{}l cccccc@{}}

\toprule

{} & PPL & F1 & Avg. & Ext. & Greedy\\

\midrule

ITDD~\citep{Li2019IncrementalTW}
& 26.0 & 10.4 & 0.748 & 0.390 & 0.587
\\

DRD~\citep{Zhao2020LowResourceKD}
& 46.1 & 10.8 & 0.791 & 0.406 & 0.613
\\

\midrule

TMN~\citep{Dinan2019WizardOW}
& 75.2 & 9.9 & 0.789 & 0.399 & 0.615
\\

KGPT~\citep{Zhao2020KnowledgeGroundedDG}
& 20.6 & 13.5 & 0.837 & 0.437 & 0.654
\\

\midrule

% \textbf{GenKS 2SG} 
% & 32.7 & \textbf{22.3} & \textbf{38.0} & 27.9
% \\

\textbf{GenKS} 
& \textbf{16.1} & \textbf{14.1} & \textbf{0.847} & \textbf{0.442} & \textbf{0.668}
\\
%  0.812 0.430 0.641
\bottomrule
\end{tabular}
\caption{Results on CMU\_DoG in term of response quality. The best results are highlighted with \textbf{bold}.}
\label{table:gen-cmu}
\end{table}

%% file: tables/case.tex
\begin{table}[!t]
\small \centering
\setlength\tabcolsep{2pt}
\begin{tabular}{@{}rp{6.5cm}@{}}

\toprule
\textbf{Topic} & Nickelback\\
\textbf{User} & Do you like Nickelback?\\
...\\
\textbf{User} & Yes, what more can you tell me about Nickelback?\\
\textbf{System} & Chad Kroeger is the leading vocalist of the group.\\
\textbf{User} & Amazing. What about the other group members\\
\midrule
\textbf{Passage} & The band is composed of guitarist and lead vocalist \emph{Chad Kroeger}, guitarist, keyboardist and backing vocalist \emph{Ryan Peake}, bassist \emph{Mike Kroeger}, and drummer \emph{Daniel Adair}.\\
\midrule
\textbf{TMN} &{i think the song is a very popular song}\\
\textbf{DukeNet} &{chad kroeger is a great band}\\
\textbf{KGPT} &{the lead vocalist is chad kroeger. he is also the drummer.}\\
\textbf{GenKS} & Chad Kroeger, Ryan Peake, Mike Kroeger and Daniel Adair.\\
\textbf{Human} & Ryan Peake is the keyboardist an Mike Kroeger is the bassist.\\
\bottomrule
\end{tabular}
\caption{Case study on the Wizard Test Unseen dataset. This table shows an example where both \textsc{GenKS} and baselines select the proper knowledge.} \label{table:case1}
\end{table}

% ============================

\begin{table}[!t]
\small \centering
\setlength\tabcolsep{2pt}
\begin{tabular}{@{}rp{6.5cm}@{}}

\toprule
\textbf{Topic} & List of national parks of the United States\\
\textbf{User} & I love national parks dont you\\
...\\
\textbf{User} & I live in Arkansas and love the Hot SPrings National Park and have been there many times, really it is beautiful\\
\midrule
\textbf{DukeNet} &{i have heard the city is located in the ouachita mountains among the us}\\
\textbf{KGPT} &{i've been to the ouachita mountains, too! i've been to the ouachita mountains in the ouachita mountains.}\\
\textbf{GenKS} & I've never been to Hot Springs, but I’ve always wanted to go to there.\\
\textbf{Human} & I've never been to that one! I bet its beautiful!\\
\bottomrule
\end{tabular}
\caption{Case study on the Wizard Test Unseen dataset. This table shows an example where both models select the wrong knowledge.} \label{table:case2}
\end{table}